\documentclass[letterpaper, 10 pt, conference]{ieeeconf}  

\IEEEoverridecommandlockouts                              

\usepackage{algorithm}
\usepackage{algpseudocode}
\usepackage{adjustbox}
\usepackage{array}
\usepackage{booktabs}

\begin{document}

\title{\LARGE \bf
Meta-aware Learning in text-to-SQL Large Language Model*}

\author{Wenda Zhang$^{1}$\thanks{$^{*}$This work was supported by Walmart Global Tech}\thanks{$^{1}$Wenda Zhang is with Walmart Global Tech, Sunnyvale, CA, USA {\tt\small zhangwenda1990@gmail.com}}}

\maketitle
\thispagestyle{empty}
\pagestyle{empty}

\begin{abstract}

The advancements of Large language models (LLMs) have provided great opportunities to text-to-SQL tasks to overcome the main challenges to understand complex domain information and complex database structures in business applications. In this paper, we propose a meta-aware learning framework to integrate domain knowledge, database schema, chain-of-thought reasoning processes, and metadata relationships to improve the SQL generation quality. The proposed framework includes four learning strategies: schema-based learning, Chain-of-Thought (CoT) learning, knowledge-enhanced learning, and key information tokenization. This approach provides a comprehensive understanding of database structure and metadata information towards LLM through fine-tuning to improve its performance on SQL generation within business domains. Through two experimental studies, we have demonstrated the superiority of the proposed methods in execution accuracy, multi-task SQL generation capability, and reduction of catastrophic forgetting. 

\end{abstract}

\setcounter{footnote}{1}
\section{INTRODUCTION}

\label{intro}

As one of the most important tasks in natural language processing (NLP) to translate plain human language into structured query language (SQL), text-to-SQL enables non-experts to interact effectively with data in the increasingly complex business database environments \cite{hong2024next}. The recent advancements of large language models (LLMs) have presented great opportunities to the progress of text-to-SQL models. The LLM-based text-to-SQL approaches including in-context learning (ICL) \cite{pourreza2024din} and fine-tuning (FT) \cite{li2024codes} offer significant advantages over traditional rule-based approaches \cite{li2014constructing, mahmud2015rule} and neural network based approaches \cite{sutskever2014sequence, vaswani2017attention} by providing flexibility in handling complex databases, improving adaptability to a variety of systems, and saving development efforts in SQL generation tasks.

\begin{figure}[!htb]
  \begin{center}
  \includegraphics[width=0.8\linewidth]{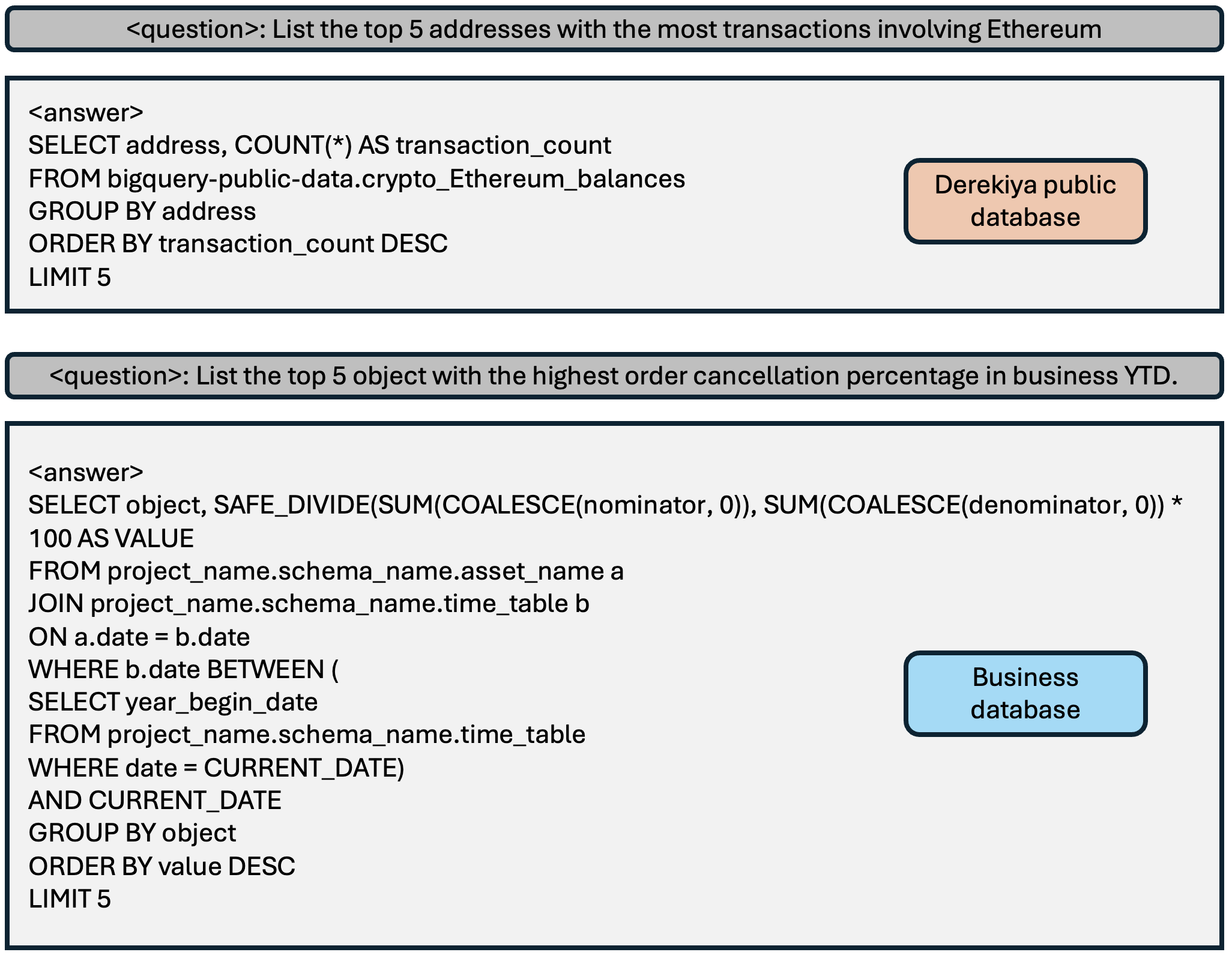}
  \caption{The comparison in SQL complexity from public database and the business database. In the question from business database, YTD refers to ``year to date" and the definition of years differs from the standard calendar year. The date information is maintained across a separate table. The key information is masked to present the SQL structures. }
  \label{figure: sql example}
  \end{center}
\end{figure}

However, translating natural language to SQL in real-world business contexts presents several challenges. Business databases often exhibit complexity, including complex schemas, domain-specific topics and data formats, and platform-specific dialects such as BigQuery SQL. In-context learning with zero-shot methods \cite{rajkumar2022evaluating} on existing text-to-SQL LLMs trained on public SQL databases (d.g., SQLCoder \footnote{https://defog.ai/blog/open-sourcing-sqlcoder}, DAIL-SQL \cite{gao2023text} and GPT-4 \cite{achiam2023gpt}) face difficulties in identifying the domain-specific unique complexities and knowledge inherent in business databases. An example of SQL structure is shown in Figure~\ref{figure: sql example}, two similar questions with the topic of ``top 5 object" are listed from a BigQuery public database \footnote{https://huggingface.co/datasets/derekiya/BigQuery\_dataset-v3} and a business database in BigQuery dialect. The business database uses non-standard date definitions and maintains date information across separate tables.

Additionally, few-shot learning requires large sets of example queries to derive the desired output. As database complexity increases, the demand for input tokens grows exponentially. This may lead to significantly increased associated costs on proprietary LLMs or encounter limitations in input tokens or devices on open-source LLMs \cite{hong2024next}. These issues can impede LLMs from effectively incorporating the essential business domain-specific information and exacerbate issues related to hallucinations \cite{zhang2023siren}. Fine-tuning an open-source LLM for specific business environments on text-to-SQL tasks is a commonly adopted solution \cite{hong2024next, li2024codes}, which can partially mitigate these issues. However, a vanilla fine-tuning \cite{sun2023sql, zhang2024benchmarking} may not effectively incorporate complex domain knowledge including specialized terms, abbreviations, custom time-frame definitions, and complex table relationships.

To address these challenges, we propose a meta-aware learning strategy designed to enable LLMs to incorporate domain knowledge through fine-tuning, optimizing their performance for domain-specific, complex business database environment. By concentrating on multiple aspects in a business domain including table schema details, chain-of-thought reasoning process to generate SQL for a business question, domain knowledge relationships, and key information integration, the proposed method can increase the capability for an LLM to generate accurate SQL queries in a complex business database environment.


The paper is structured as follows: Section \ref{sec related work} presents the related work to this study. Section \ref{sec: methodology} outlines the proposed meta-aware learning approach, elaborating the components including schema-based learning, chain-of-thought (CoT) learning, domain knowledge enhancement learning, and key information tokenization method. In section \ref{sec experiments}, we present experiment results on real business databases at Walmart to evaluate the effectiveness of our approach on multi-table domain-specific tasks, followed by conclusions and future work directions in section \ref{sec conclusion}.

\section{RELATED WORK}
\label{sec related work}

Research on text-to-SQL has been greatly developed in recent decades. The early work of text-to-SQL research was conducted with rule-based approaches, which aimed to parse information with natural language processing (NLP) techniques from input questions to generate SQL queries using predefined templates and syntax rules \cite{li2014constructing, mahmud2015rule}. 

Recent advancements in deep learning have significantly facilitated the development of syntax parsing by integrating the mapping function from input natural languages to the SQL output \cite{guo2019towards, li2023resdsql, gao2023text}. Deep learning methods, including long-short-term memory (LSTM) \cite{guo2019towards}, convolutional neural network (CNN) \cite{o2015introduction}, sequence-to-sequence (Seq2Seq) \cite{sutskever2014sequence}, and the attention mechanism \cite{vaswani2017attention}, have been used in questions and database information encoding to improve context integration and syntax understanding in text-to-SQL tasks.

The emergence of large language models (LLMs) has provided a groundbreaking solution to text-to-SQL \cite{hong2024next, gao2023text}, with significant capability in syntax understanding, fluent output generation, and knowledge storage. Pre-trained LLMs, such as GPT \cite{achiam2023gpt}, with in-context learning (ICL) \cite{dong2022survey, sahoo2024systematic}, can effectively synthesize schema information from input prompts to generate accurate SQLs \cite{rajkumar2022evaluating, yoon2018dynamic}. By providing reasoning steps from input questions to formulate the SQL output, the Chain-of-thought (CoT) approach \cite{wei2022chain, zhang2023act, tai2023exploring} can enhance the performance of SQL generation. Furthermore, retrieval-augmented generation (RAG) \cite{lewis2020retrieval, thorpe2024dubo} is a promising technique recently used to enhance text-to-SQL performance by integrating relevant context from external sources and improving understanding of domain-specific queries.

Another widely adopted approach to leverage LLM's capability is to integrate question-SQL mapping information through fine-tuning techniques \cite{hong2024next, li2024codes, sun2023sql, zhang2024benchmarking}. However, the implementation of fine-tuning in text-to-SQL LLMs is prone to issues like catastrophic forgetting \cite{shi2024survey, fernando2024mitigating}, where pre-trained knowledge is lost during fine-tuning, low computational efficiency \cite{shi2024survey}, and hallucination \cite{zhang2023siren}, where models generate inaccurate or unrelated SQL queries to the provided schemas and contexts.

To address these issues, researchers have explored strategies in LLM fine-tuning for text-to-SQL applications, including parameter-efficient fine-tuning (PEFT) methods \cite{li2024codes, hu2021lora}, domain-specific knowledge incorporation \cite{hong2024next}, and multi-task fine-tuning strategies \cite{mahabadi2021parameter}, which provided novel perspectives on text-to-SQL solutions.

\section{METHODOLOGY}
\label{sec: methodology}
In this paper, we propose a meta-aware learning framework to improve text-to-SQL performance by integrating multiple targeted learning strategies. This framework combines several complementary subprocesses: schema-based learning, chain-of-thought reasoning, domain knowledge enhancement, and key information tokenization. Each component is designed to address specific challenges in text-to-SQL tasks, such as schema comprehension, logical reasoning, domain-specific adaptation, and critical token identification. In this section, we provide a detailed examination of each component and its role in the meata-aware learning framework.


\begin{figure}[!htb]
  \begin{center}
  \includegraphics[width=1\linewidth]{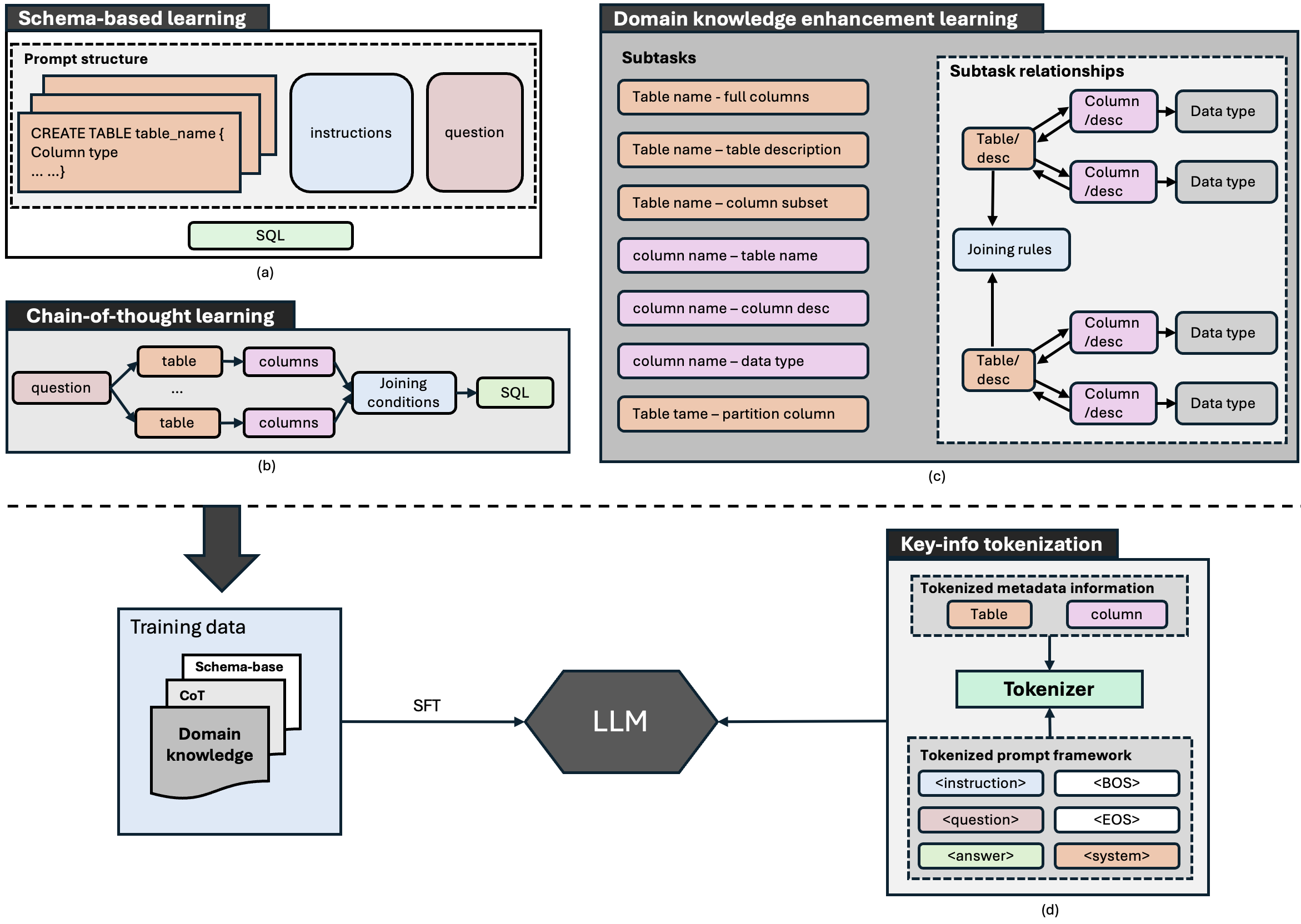}
  \caption{The framework of meta-aware learning with sub-modules: (a) schema-based learning module with prompt structure; (b) Chain-of-Thought (CoT) learning module with step-by-step reasoning process; (c) domain knowledge enhancement learning module with sub-tasks and relationships; (d) key information tokenization module with tokenized elements.}
  \label{figure: meta-aware framework}
  \end{center}
\end{figure}

\subsection{Schema-based learning}
\label{subsec: schema}
As shown in Figure~\ref{figure: meta-aware framework}.a, schema-based learning, as the most popular learning method for text-to-SQL LLM fine-tuning, refers to a learning paradigm in which the structure and metadata of a database schema are directly integrated into the model's learning process to guide text-to-SQL query generation \cite{zhong2017seq2sql}. In our proposed method, we include the exact SQL-related tables and columns in the schema-based learning strategy.

\subsection{Chain-of-thought learning}
\label{subsec: COT}
Chain-of-Thought (CoT) learning in the context of text-to-SQL refers to a reasoning-based approach where the model generates intermediate reasoning steps before producing the final SQL query \cite{wei2022chain}. As presented in Figure~\ref{figure: meta-aware framework}.b, CoT learning breaks down the query generation process into a series of logical sub-tasks, including identifying target tables, columns, relationships between tables, and generating SQL outputs.

\subsection{Domain knowledge enhancement learning}
\label{subsec: domain enhancement}
Domain knowledge enhancement learning refers to the process of integrating domain-specific metadata into text-to-SQL LLMs, which aims at enriching the model’s knowledge with database-specific details, including table and column names, descriptions of attributes, and relationships between tables. This strategy is designed to align LLM with the specific structure of the target database. In this paper, we incorporated comprehensive aspects of domain knowledge into LLM with seven subtasks as shown in Figure~\ref{figure: meta-aware framework}.c: (1) identifying relevant columns from given tables; (2) recognizing table names based on provided descriptions; (3) identifying tables corresponding to given columns; (4) generating descriptions for tables; (5) generating descriptions for columns; (6) identifying the data type of a given column; and (7) understanding relationships between tables to handle JOIN operations. 

\subsection{Key information tokenization}
\label{subsec: key info token}

The key information tokenization contains two components: metadata tokenization and prompt framework tokenization (Figure~\ref{figure: meta-aware framework}.d).

Metadata tokenization involves the systematic conversion of database table and column names into tokens and the integration with the model's tokenizer. This process is designed to establish a robust association between the large language model (LLM) and the underlying database schema, enhancing the model's capacity to interpret and interact with the data structures effectively \cite{czapla2018universal}.

\begin{table*}[!htb]
    \centering
    \caption{Prompt structures used for training in different strategies}
    \resizebox{\textwidth}{!}{\begin{tabular}{@{}>{\raggedright\arraybackslash}p{3cm} 
                    >{\raggedright\arraybackslash}p{6cm} 
                    >{\raggedright\arraybackslash}p{5cm}
                    >{\raggedright\arraybackslash}p{5cm}@{}}
        \toprule

        \multicolumn{4}{c}{\textbf{Meta-aware Prompt}}\\
        \midrule
        \textbf{ } & \textbf{(a) Schema-based learning} & \textbf{(b) CoT learning} & \textbf{(c) knowledge learning}\\
        \cmidrule(lr){2-4}
        \texttt{\newline 
        \newline
        \newline
        \newline
        \newline
        \newline
        \newline
        \newline
        Prompt Structure} &
        \texttt{<s> \newline <system> \newline Assistant claimer: SQL expert \newline Schema: CREATE TABLE ( col1 datatype desc \newline col2 datatype desc \newline ...) \newline <end> \newline <instruction> \newline 1. Joining rules \newline 2. default values \newline <end> \newline <question> Question \newline <end> \newline <answer> Answer \newline <end> \newline </s>} & 
        \texttt{<s> \newline <system> \newline Assistant claimer: CoT SQL expert \newline <end> \newline <instruction> \newline 1. identify the related tables \newline 2. identify the columns used to solve the question \newline 3. identify the columns to join the tables \newline 4. generate SQL to solve the question \newline <end> \newline <question> Question \newline <end> \newline <answer> Answer \newline <end> \newline </s>} & 
        \texttt{<s> \newline 
        <system> \newline 
        Assistant claimer: metadata knowledge assistant \newline 
        <end> \newline 
        <question> Question \newline 
        <end> \newline 
        <answer> Answer \newline 
        <end> \newline 
        </s>}\\ 
        \midrule
        \multicolumn{4}{c}{\textbf{Base Prompt}}\\
        \midrule
        \textbf{} & \textbf{(d) Base-prompt-I} & \textbf{(e) Base-prompt-II} \\ 
        \cmidrule(lr){2-3}
        \texttt{\newline 
        \newline
        \newline
        \newline
        \newline
        \newline
        Prompt Structure} & 
         
        \texttt{<s>[INST] \newline <<SYS>> \newline Assistant claimer: SQL expert \newline Schema: CREATE TABLE ( col1 datatype desc \newline col2 datatype desc \newline ...) \newline <</SYS>> \newline Instructions: \newline 1. Joining rules \newline 2. default values \newline Question \newline [/INST] \newline Answer \newline </s>} & \texttt{[Schema] CREATE TABLE ( col1 datatype desc \newline col2 datatype desc \newline ...) \newline [Question] Question \newline [Answer] Answer} &\\ 
        
        \bottomrule
    \end{tabular}}
    \label{tab:prompt_structures}
\end{table*}

Prompt framework tokenization refers to the structuring of prompts with tokenized tags, which serve to clarify the task requirements and enhance the model's comprehension of the specific objectives, facilitating task-specific understanding and performance. The proposed prompt framework is shown in Table~\ref{tab:prompt_structures}.a, Table~\ref{tab:prompt_structures}.b, and Table~\ref{tab:prompt_structures}.c

\section{EXPERIMENTS}
\label{sec experiments}
We performed several experimental studies to prove the effectiveness of the aforementioned strategies. Two scenarios were carried out structured tagging system
Figure~\ref{figure: meta-aware framework} shows the framework of the proposed meta-aware learning method and the details of each learning method. Due to Walmart Privacy Requirements, models and datasets are not open to public.

\subsection{Datasets}
\label{subsec dataset}
We prepared two experimental datasets to study the performance and effectiveness of the proposed learning methods in domain-specific business areas. The datasets were generated from tables in business databases.

We employed template-based approaches \cite{yu2018typesql, yu2020grappa} in our data generation process. As presented in Algorithm~\ref{alg:generate_dataset}, We first prepared a set of golden question-SQL pairs\cite{yu2018typesql}, $\{Q, A | m\in M, f\in F\}$, where $Q$ denotes the question and $A$ refers to the corresponding SQL, as templates that comprehensively cover topics from the targeted domain. Let $m\in M$ be the metrics and $f\in F$ be the values of filters required in formulating the question-SQL pairs. These values $\{m, f\}$ were then generated from the metadata in the domain database to derive the full set of question-SQL pairs. 

The prompt set $P = \{S, I, Q, A| T\}$ was then generated using system information ($S$) which refers to the LLM assistant claimer and schemas in this paper, instructions ($I$), and question-SQL pairs $\{Q, A\}$ with the given prompt structure ($T$).

The training of LLM requires diversified prompt to avoid overfitting or severe bias \cite{hernandez2018data}. We employed an LLM-based diversification method \cite{cui2023ada} to introduce variation in the prompt by creating modified versions of the original instruction. Considering the computational complexity to implement an LLM-based diversification method on a large training set, a random sampling technique was employed to replace the instructions in the set of prompt with sampled instructions from a smaller set of preliminarily diversified instructions ($\mathcal{I}$).

\begin{algorithm}
\caption{Training set preparation}
\label{alg:generate_dataset}
\begin{algorithmic}[1]
\Require Initial prompt set $P_i = \{S, I, Q_i, A_i| T\}$ with template question-SQL pairs $\{Q_i, A_i | m_i, f_i\}$, $i = 1,...,l$ where $l$ is the number of question-SQL pairs in the template set. $S, I, T$ are static system information, instruction and the prompt structure; and $m_i, f_i$ are the metrics and filters in the template for the $i$-th sample. 
\State \textbf{Step 1:}
\For{$i = 1,...,l$}
    \State (I) Generate $\{Q_i, A_i\} = \{Q_i, A_i | m_i, f_i\} \times \{m_i \in M\} \times \{f_i\in F\}$.
    \State (II) Derive the prompt $P_i = \{S, I, Q_i, A_i| T\}$ for each $\{Q_i, A_i\}$.
\EndFor
\State Let $|P|,|M|,|F|$ denote the size of $P, M, F$; and $|P| = l\cdot |M|\cdot|F|$.
\State \textbf{Step 2:}
\For{$k = 1,...,K$, where $K << |P|$}
    \State $I_k$ $\gets$ rephrase $I$ with LLM-based diversification method
\EndFor
\State \textbf{Step 3:}

\For{$j = 1,...,J$, where $J = |P|$}
    \State (I) Sample $I_j \in \mathcal{I}$, where $|\mathcal{I}| = K$.
    \State (II) $P_j = \{S, I_j, Q_j, A_j| T\} \gets I_j$, replace the static $I$ with $I_j$ sampled from $\mathcal{I}$.
    \State (III) Append $P_j$ to training set $\mathcal{D}$
\EndFor

\end{algorithmic}
\end{algorithm}

\subsubsection{Scenario I}
\label{subsub: scenario1}
In scenario I, we aimed to assess the performance of the prompt structure tokenization method, where we provided a list of structure tags, such as $\texttt{<system>}$ / $\texttt{<instruction>}$ / $\texttt{<question>}$ / $\texttt{<answer>}$, in the tokenizer to structure the prompt. To control the effect of table variation, the dataset was generated from a single domain table which contains $<$10 filter columns, $<$20 business calculation metrics, and $<$200 value of filters, resulting in a dataset size of $\approx$30,000 entries. We sampled 500 data points in the testing dataset and created multiple training subsets of varying sizes ($n = 250, 500, 750, 1000, 5000, 10000$). We compared the prompt structure tokenization method with the base prompt training method with base prompt I as shown in Table~\ref{tab:prompt_structures}.d.

\subsubsection{Scenario II}
\label{subsub: scenario2}
In scenario II, the dataset was generated from 5 tables which contains $<$80 filter columns, $<$20 metrics, and $<$200 values of filters in each table, which results in a full dataset of $\approx$1,000,000 entries. In this scenario, we aimed to evaluate the model's capability for cross-table SQL generation and the effectiveness of the proposed meta-aware learning method through fine-tuning. We averagely sampled 500 data from the 5 tables to acquire a balanced testing set. The training set with a size of 5,000 was also sampled averagely from the full dataset.

\subsection{Evaluation metrics}
\label{subsec: evaluation}
We use the execution accuracy \cite{hong2024next, gao2023text} to evaluate the validity of the SQL output from the LLM in both scenario I and II, which compares the execution output of the predicted SQL and the ground truth SQL queries from the business database \cite{yu2018spider}. In scenario I, steps to overfitting and the training time were also measured to present the performance of the proposed methods.

\subsection{Competing methods}
\label{subsub competing}

We consider several competing methods with various combinations of schema-based learning (Schema), Chain-of-Thought learning (CoT), domain knowledge enhancement learning (Kn), and key information tokenization (OPT) to evaluate the performance of the proposed methods:

\begin{itemize}  
    \item \textbf{Schema (base-prompt-II)}: This baseline method utilizes only schema-based information with structure-free training data, as demonstrated in Table~\ref{tab:prompt_structures}.e.
    \item \textbf{Schema-CoT-Kn}: This method integrates schema-based learning, Chain-of-Thought learning, and domain knowledge enhancement learning, enabling the model to enhance its understanding of the data from multiple perspectives, thereby strengthening its metadata knowledge.
    \item  \textbf{Schema-CoT-OPT}: This approach combines schema-based learning and Chain-of-Thought learning with key information tokenization, which tokenizes metadata information during the fine-tuning process.
    \item \textbf{Schema-CoT-Kn-OPT}: This method merges the Schema-CoT-Kn approach with key information tokenization.
    \item \textbf{Schema (base-prompt-I)}: In this method, schema-based learning is conducted solely with the basic prompt structure \cite{roziere2023code, sinha2024pae}, as illustrated in Table~\ref{tab:prompt_structures}.d.
    \item \textbf{Kn@Schema/Schema@Kn}: These two methods represent a step-by-step progressive learning approach, where a fine-tuned model is initially developed on one dataset and subsequently refined into the final model based on the prior model. Kn@Schema indicates that domain knowledge enhancement (Kn) is performed first, followed by schema-based learning, while Schema@Kn denotes the reverse order.
    \item \textbf{Kn}: This method focuses exclusively on domain knowledge enhancement learning for the large language model (LLM), allowing it to learn metadata information without the SQL-question pair context.
\end{itemize}

\subsection{Parameters}
\label{subsec: parameter}

Our fine-tuning experiments utilize the open-source LLM model (models are not open to public due to Walmart privacy requirements). Key training hyper-parameters are set as follows:
\begin{itemize}
    \item Optimization: We use the AdamW optimizer with a learning rate of $2\times10^{-5}$, a dropout rate of 0.05, and a weight decay rate of 0.003 to help mitigate overfitting.
    \item LoRA parameters: We use low-rank adaptation (LoRA) as the fine-tuning method \cite{hu2021lora}, with a scaling factor $\alpha=64$ and the rank $r=32$ to ensure a reasonable scale of the weight matrix \cite{song2024increasing} and fine-tune the q\_proj, k\_proj, v\_proj, and o\_proj layers.
    \item Batch size: Training is performed with a per-device batch size of 1 on a single V100 32GB GPU. To accommodate larger effective batch sizes without exceeding memory limits, we employ a gradient accumulation strategy with an accumulation step of 1.
    \item Training steps and evaluation: We cap our training steps at 4000, using an early stopping criterion to detect overfitting. Evaluation is conducted every 200 steps to monitor performance, with the stopping step recorded for overfitting analysis.
    \item Precision: To strike a balance between computational efficiency and precision, we use the bfloat16 format throughout training, enhancing overall performance while conserving resources.
    \item Inference parameters: We use non-deterministic sampling (do\_sample=True) with 2 independent outputs (num\_beams=2) for beam search method in the inference process to generate SQL output.
\end{itemize}
This parameter configuration is consistently applied across all experimental scenarios.

\subsection{Results}
\label{subsec: results}

\textit{Scenario I.} Table~\ref{table1: multi-task performance} compares the proposed prompt framework and the basic prompt structure across various sample sizes ($n=250, 500, 750, 1000, 5000, 10000$) in terms of execution accuracy, steps to overfitting, and training time. The tokenized prompt method consistently outperformed the base prompt method across all sample sizes, with the performance difference more pronounced at smaller sample sizes. With 250 training samples, the tokenized prompt method achieved an accuracy of 0.914, 27.7\% higher than that of the base prompt method. We observed that $n=1,000$ was sufficient to achieve the best accuracy with a single-table schema.

\begin{table*}[!htb]
    \centering
    \caption{Performance comparison of tokenized- and base-prompt training strategies across training set sample sizes ($n$). The values in parentheses represent the percentage difference of the tokenized prompt training strategy compared to the base prompt training strategy. Training times are measured in hours.}
    \resizebox{\textwidth}{!}{%
        \begin{tabular}{cccc|ccc}
            \toprule
            \textbf{} & \multicolumn{3}{c}{\textbf{Schema-based learning}} & \multicolumn{3}{c}{\textbf{Schema (base-prompt-I)}} \\
            \cmidrule(lr){2-4} \cmidrule(lr){5-7}
             \textbf{Training set} & \textbf{Steps to} & \textbf{Execution} & \textbf{Training} & \textbf{Steps to} & \textbf{Execution} & \textbf{Training} \\
             \textbf{sample size(n)} & \textbf{overfitting} & \textbf{accuracy} & \textbf{time (h)} & \textbf{overfitting} & \textbf{accuracy} & \textbf{time (h)} \\
            \midrule
            \textbf{n=250}   & 1200 & 0.914 (+27.7\%) & 6.3 (+42.4\%) & 600  & 0.716 & 4.4 \\
            \textbf{n=500}   & 2200 & 0.960 (+8.4\%) & 4.4 (-1.2\%) & 1200 & 0.886 & 4.5 \\
            \textbf{n=750}   & 2800 & 0.918 (+8.8\%) & 6.5 (+19.0\%) & 1600 & 0.844 & 5.5 \\
            \textbf{n=1,000}  & 3800 & 0.962 (+3.9\%) & 6.2 (+10.7\%) & 3200 & 0.926 & 5.6 \\
            \textbf{n=5,000}  & 4000 & 0.966 (+1.5\%) & 7.2 (+6.1\%) & 4000 & 0.952 & 6.8 \\
            \textbf{n=10,000} & 4000 & 0.964 (+1.0\%) & 10.3 (+7.6\%) & 4000 & 0.954 & 9.5 \\
            \bottomrule
        \end{tabular}%
    }
\label{table1: multi-task performance}
\end{table*}

\begin{figure}[H]
  \begin{center}
  \includegraphics[width=0.8\linewidth]{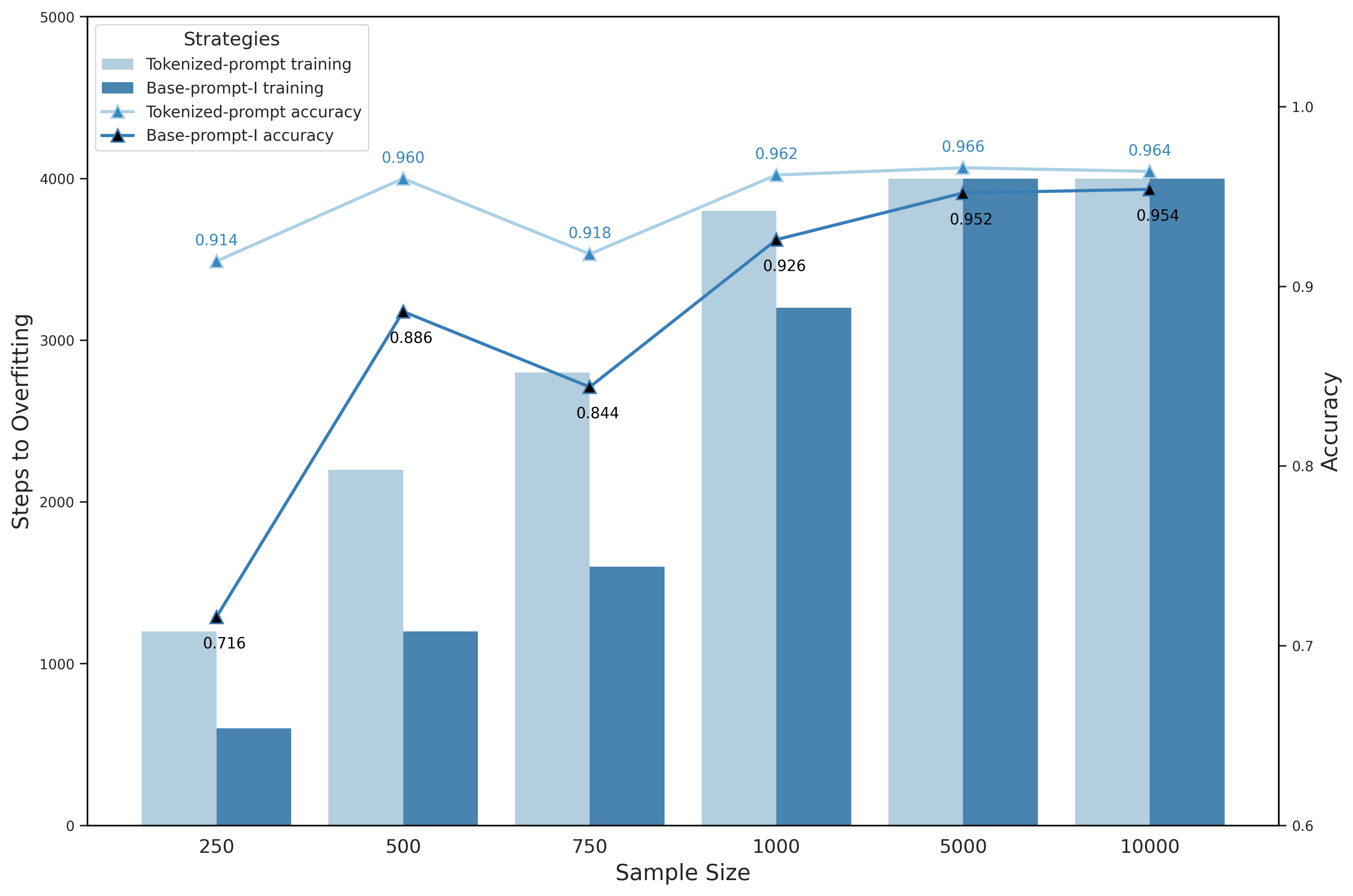}
  \caption{Performance comparison plots between tokenized-prompt training (prompt structure tokenization) and based-prompt-I (Table~\ref{tab:prompt_structures}.d) training strategies. Bar plot shows the steps to observe overfitting across different training set sample sizes. The line plots present the accuracy comparison of the two strategies.}
  \label{figure: barplot}
  \end{center}
\end{figure}

Although the tokenized approach required, on average, 9.9\% longer training time than the base prompt, this increase is due to the prompt structure tokenization step, which expands the embedding size and thus increases computational complexity. Despite this added cost, the tokenized prompt method offers clear advantages: it consistently delayed overfitting compared to the base prompt, as shown in Figure~\ref{figure: barplot}, enhancing model stability and robustness. These benefits make the tokenized approach preferable, especially for smaller datasets where the structured, tokenized inputs help maintain accuracy and reduce overfitting tendencies.

\textit{Scenario II.} In Scenario II, the model was trained using the exact schema, containing only the tables and columns referenced in each ground truth SQL query. To simulate real-world applications where the exact schema is often unknown, we tested the model with two schema formats: a full schema, containing all available tables and columns, and a dynamic schema, a subset of tables and columns relevant to the schemas used in the training data. For CoT testing cases, we prepared step-by-step reasoning instructions and questions using the CoT prompt structure described in \ref{subsec: COT}.

As demonstrated in Table~\ref{table2: single-task performance}, meta-aware learning approaches achieved the highest execution accuracy across both full and dynamic schema testing cases, with Schema-CoT-OPT reaching top scores of 0.928 and 0.930, respectively. Models trained solely on schema-based data underperformed on schema-specific tasks, with accuracies below 0.7. We observed that CoT instructions enabled schema-only models to achieve accuracies above 0.9 in CoT tasks. Among all methods, the meta-aware learning approach incorporating domain knowledge—specifically the Schema-CoT-Kn method—achieved the highest execution accuracy of 0.940 in CoT tasks.

\begin{table}[!htb]

\centering
\caption{Model Performance Comparison}
\begin{tabular}{lccc|c}
\hline
\addlinespace[0.3em]
 & \textbf{Full} & \textbf{Dynamic} &  & \textbf{Base-} \\ 
\textbf{Method} & \textbf{Schema} & \textbf{Schema} & \textbf{CoT} & \textbf{ Prompt-I} \\ 
\addlinespace[0.3em]
\hline
\addlinespace[0.3em]
Schema (base-prompt-II) & $<$0.1 & $<$0.1 & $<$0.1 &0.123\\ \hline
\addlinespace[0.3em]
\multicolumn{5}{c}{\textbf{Meta-aware Learning}}\\ 
\addlinespace[0.3em] 
\hline
\addlinespace[0.3em]
Schema-CoT-Kn & 0.918 & 0.905 & \textbf{0.940} & 0.878 \\ 
Schema-CoT-OPT & \textbf{0.928} & \textbf{0.930} & 0.925 & \textbf{0.928} \\ 
Schema-CoT-Kn-OPT & 0.923 & 0.892 & 0.892 & 0.913 \\ 
\addlinespace[0.3em]
\hline
\addlinespace[0.3em]
\multicolumn{5}{c}{\textbf{Progressive Learning}}\\
\addlinespace[0.3em]
\hline
\addlinespace[0.3em]
Schema & 0.635 & 0.688 & 0.908 & 0.482 \\ 
Kn@Schema & 0.513 & 0.545 & 0.915 & 0.418 \\ 
Schema@Kn & $<$0.1 & $<$0.1 & $<$0.1 & $<$0.1 \\ 
Kn & $<$0.1 & $<$0.1 & $<$0.1 & $<$0.1 \\ 
\addlinespace[0.3em]
\hline
\end{tabular}
\label{table2: single-task performance}
\end{table}

The schema-only learning strategy with base-prompt-II as shown in Table~\ref{tab:prompt_structures}.e failed to capture the complex structures in a real-world business database environment, with accuracy in all testing cases lower than 0.123. 

Progressive learning \cite{fayek2020progressive} is a strategy for sequentially learning tasks while retaining knowledge from prior tasks in multi-task learning. However, progressive learning can be prone to catastrophic forgetting \cite{fayek2020progressive, zhai2024investigating}, where knowledge from previous tasks is lost as weights are updated for subsequent tasks. The comparison among Schema, Schema@Kn, and Kn@Schema presents the significant impact of catastrophic forgetting \cite{zhai2024investigating} on text-to-SQL LLM fine-tuning tasks. As shown in Table~\ref{table1: multi-task performance}, introducing Kn after Schema-based learning masked prior text-to-SQL knowledge, reducing all text-to-SQL task accuracies to below 0.1. Additionally, treating Schema-based learning in Kn@Schema as the final task in progressive learning still reduced schema-based SQL generation accuracy by 10–20\%.

The proposed meta-aware learning method demonstrated superior performance compared to the based line and other learning methods. Among the meta-aware approaches, Schema-CoT-OPT and Schema-CoT-Kn showed the best results, indicating that integrating Chain-of-Thought learning and domain knowledge with schema-based learning can significantly enhance the model's capabilities in text-to-SQL tasks for complex business database environment.

\section{CONCLUSIONS}
\label{sec conclusion}

This paper introduces a novel meta-aware learning method designed to enhance text-to-SQL performance in large language models (LLMs) with business database. The key contributions of this study are three-fold: (1) it presents a comprehensive framework for text-to-SQL LLM fine-tuning in complex industrial business domains; (2) it addresses the challenges of catastrophic forgetting and overfitting in multi-task text-to-SQL LLM fine-tuning; and (3) it provides an extensive comparison of diverse learning strategies, providing insights for related work and research with practical applications.  


The proposed method establishes a comprehensive framework that incorporates question-SQL mapping, database schema, SQL synthesis processes, metadata relationships, and model tokenizer to improve the LLM's ability to handle complex information in real-world business databases for text-to-SQL tasks. Experiments conducted with real-world business data in two scenarios demonstrate the effectiveness of this meta-aware learning approach in fine-tuning LLMs for text-to-SQL tasks. As shown in Section~\ref{sec experiments}, the approach significantly improves SQL generation accuracy across schema-based and reasoning contexts. Additionally, by employing structured input prompts, the method offers a clear framework that organizes the input data in alignment with the database schema, thus facilitating more accurate SQL generation and reducing the likelihood of overfitting. The structured prompts encourage the model to generalize patterns between questions and SQL outputs, rather than memorizing specific training samples, thereby supporting improved stability and performance, particularly in smaller datasets.

Our findings also emphasize the importance of effective knowledge management to prevent catastrophic forgetting, ensuring sustained high performance on text-to-SQL tasks. Moreover, the key information tokenization technique strengthens associations between the LLM and database schemas, further enhancing SQL interpretation and generation capabilities.

This work provides solutions to key challenges in text-to-SQL tasks within complex business database environments, though limitations remain regarding long-context learning and cross-domain knowledge integration. Future research could explore multi-task learning strategies with expanded input token capacities and advanced long-context schema retrieval capabilities in text-to-SQL tasks.

\addtolength{\textheight}{-12cm}   





\section*{ACKNOWLEDGMENT}

This research was supported by the Walmart. Any opinions, findings, and conclusions or recommendations expressed in this material are those of the author(s) and do not necessarily reflect the views of the funding parties.



\section*{DATA AVAILABILITY STATEMENT}
Due to Walmart's Privacy Requirements, models and datasets are not open to public.

\bibliographystyle{IEEEtran}\bibliography{mybibilo}

\end{document}